\documentclass{article}
\usepackage{spconf,amsmath,graphicx,hyperref}
\usepackage{amsmath,amssymb}
\usepackage{enumitem}

\title{Centering Emotion Hotspots: Multimodal Local-Global Fusion and Cross-Modal Alignment for Emotion Recognition in Conversations}
\name{%
\begin{tabular}{@{}c@{}}
\itshape
Yu Liu$^{\dagger,1}$ \qquad Hanlei Shi$^{\dagger,1}$ \qquad Haoxun Li$^{1}$ \qquad Yuqing Sun$^{1}$ \qquad
Yuxuan Ding$^{1}$  \\
\itshape
Linlin Gong$^{1}$ \qquad Leyuan Qu$^{\star,1}$ \qquad Taihao Li$^{\star,1}$
\end{tabular}%
\thanks{$^{\dagger}$ These authors contributed equally (co-first authors).}
\thanks{$^{\star}$ Corresponding authors.}
\noindent\thanks{Yu Liu, et al. Copyright 2026 IEEE. Personal use of this material is permitted. Permission from IEEE must be obtained for all other uses, including reprinting/republishing, creating new collective works, for resale or redistribution to servers or lists, or reuse of any copyrighted component of this work. DOI will be added upon IEEE Xplore publication.}
}
\address{$^{1}$ Hangzhou Institute for Advanced Study, University of Chinese Academy of Sciences}
\begin{document}
%
\maketitle
\begin{abstract}
Emotion Recognition in Conversations (ERC) is hard because discriminative evidence is sparse, localized, and often asynchronous across modalities. We center ERC on \emph{emotion hotspots} and present a unified model that detects per-utterance hotspots in text, audio, and video, fuses them with global features via \emph{Hotspot-Gated Fusion}, and aligns modalities using a routed \emph{Mixture-of-Aligners}; a cross-modal graph encodes conversational structure. This design focuses modeling on salient spans, mitigates misalignment, and preserves context. Experiments on standard ERC benchmarks show consistent gains over strong baselines, with ablations confirming the contributions of HGF and MoA. Our results point to a hotspot-centric view that can inform future multimodal learning, offering a new perspective on modality fusion in ERC.
\end{abstract}
\begin{keywords}
Emotion recognition in conversations, multimodal affective computing, cross-modal alignment.
\end{keywords}
\section{Introduction}
\label{sec:intro}
Emotion Recognition in Conversations (ERC)~\cite{ERCSurvey,affective} aims to identify the emotion of each utterance in a dialogue from a predefined set of categories. Recent research has advanced ERC in multiple directions, including architectural improvements and multimodal modeling. On the architecture side, early approaches predominantly used recurrent networks or Transformers to model conversational sequences~\cite{DialogueRNN}. Graph-based models~\cite{DialogueGCN} further captured long-range dependencies by representing utterances and speakers as nodes. On the multimodal side, GBAN~\cite{GBAN} employed gating to balance aligned and global features, while MMoE~\cite{yu-etal-2024-mmoe} trained independent expert modules for different interaction patterns and integrated their outputs. Across these approaches, a common practice is to rely on time-aggregated, per-modality representations at the utterance level—textual embeddings and pooled acoustic or visual descriptors—which we refer to as \emph{global features}.

\begin{figure}[!t]
\centering
\includegraphics[width=1\linewidth, trim=0 0 0 0, clip]{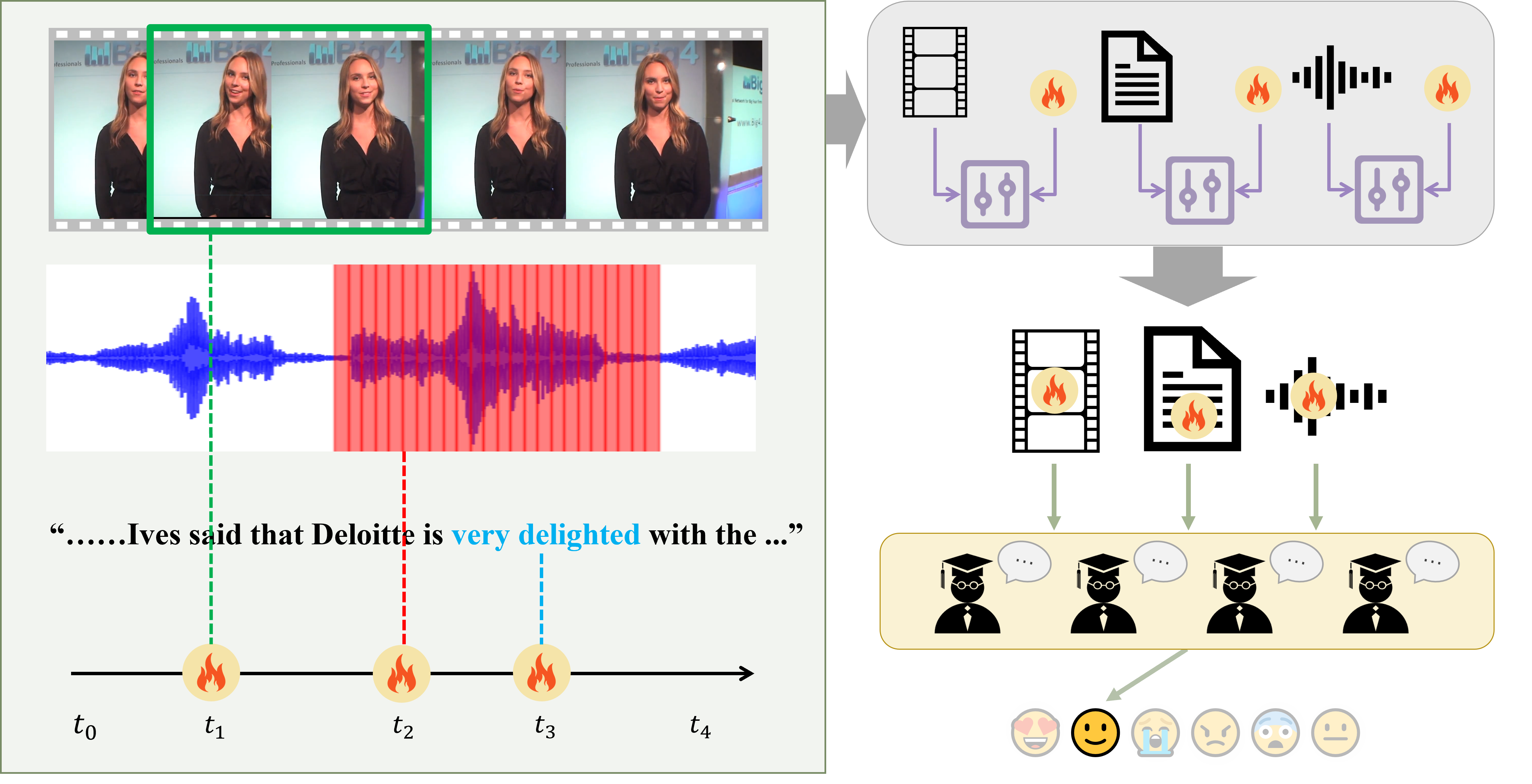}
\caption{The model locates multimodal emotion hotspots and employs gated fusion and MoA for cross-modal alignment, resolving temporal and semantic conflicts.}
\label{fig:res}
\end{figure}

Yet emotions often surface in short, high-intensity bursts: an affect-laden word in text, a brief pitch or energy spike or a laugh in audio, a fleeting eyebrow raise or wry smile in video; we identify these as \emph{emotion hotspots}. However, when tokens, clips, and frames are treated uniformly, these \emph{hotspots} are often overlooked as the abundance of neutral or weakly affective content masks scarce emotional cues. Unimodal results support the same insight: focusing on high-information \emph{hotspots} improves recognition~\cite{framekey,keyframes}. In multimodal ERC, however, these hotspots are typically \emph{asynchronous} rather than co-temporal—prosody may crest before the face reacts, and both may lead or lag the decisive word.

This asynchrony explains why global pooling or naïve alignment (uniform frame/word matching or early fusion without handling unaligned sequences) tends to dilute local evidence and underperform on misaligned data~\cite{tsai2019MULT}. To address it, we detect per-modality \emph{emotion hotspots} and softly align them around their local maxima rather than enforcing rigid time matches. We realize this with an adaptive hotspot–global fusion, a routed mixture-of-experts cross-attention for cross-modal alignment~\cite{fedus2022switch}, and a graph pathway that supplies conversational structure. The two streams are combined for utterance-level prediction. Figure~\ref{fig:res} provides an overview of the model architecture, architectural details appear in Section~\ref{sec:method}. \textbf{Our main contributions are summarized as follows:}

\begin{itemize}[leftmargin=*,noitemsep,topsep=2pt,parsep=0pt,partopsep=0pt]
    \item We introduce \textbf{Hotspot-Gated Fusion} (HGF), an adaptive, modality-agnostic mechanism that \emph{identifies and weights localized high-intensity segments (“emotion hotspots”)} within each modality and fuses them with global context.
    \item We employ HGF with \textbf{Mixture-of-Aligners} (MoA) for cross-modal alignment under temporal offsets and a conversational graph branch for dialogue structure; together they align modalities while preserving contextual dependencies.
    \item On standard ERC benchmarks, the model yields consistent performance gains over strong baselines; ablations attribute improvements primarily to HGF and MoA.
\end{itemize}
\begin{figure*}[!t]
\centering
\includegraphics[width=0.85\linewidth, trim=0 0 0 0, clip]{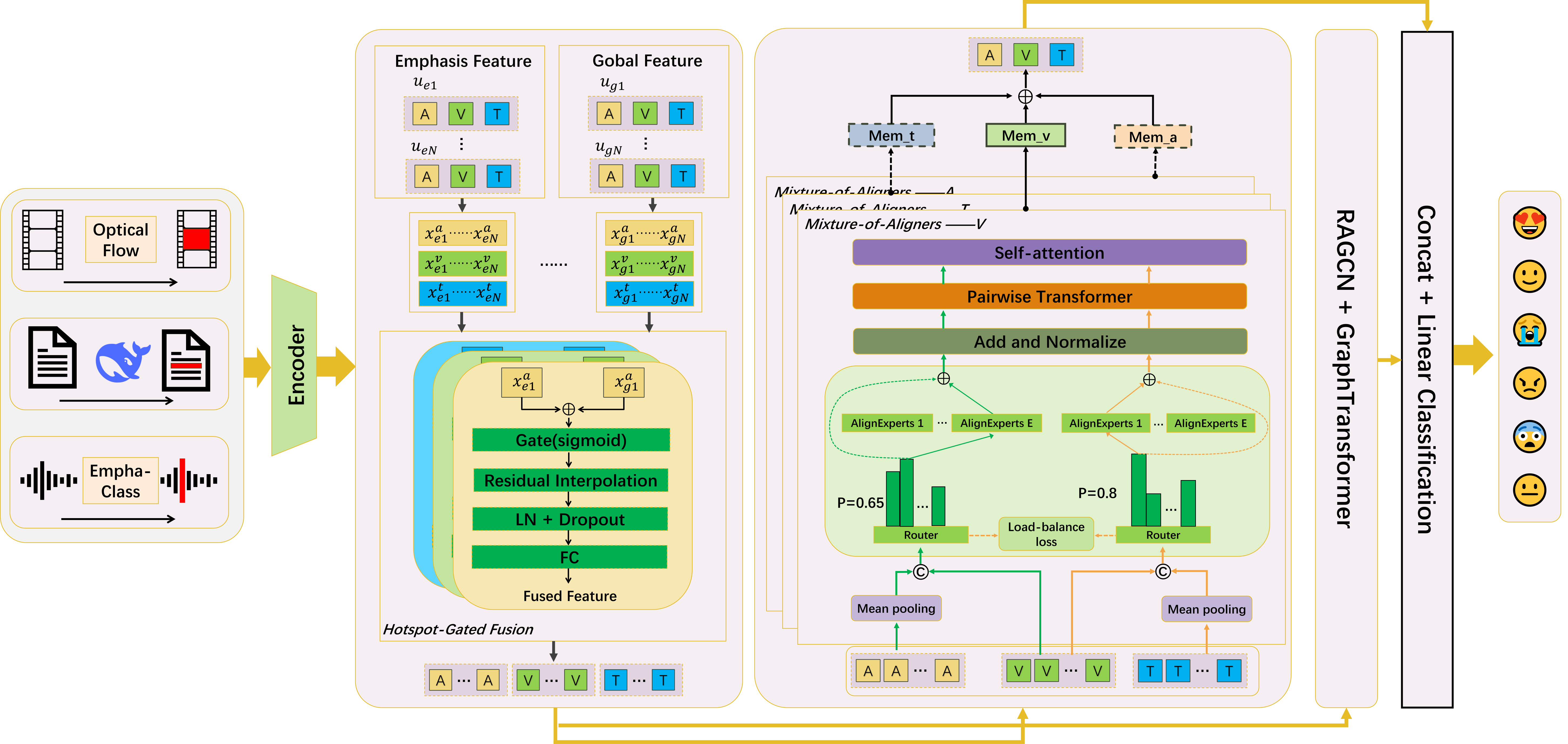}
\caption{
Per-utterance inputs from text (T), audio (A), and video (V) are first combined by \textbf{HGF} and then encoded into a shared representation.
Two pathways operate in parallel: (i) \textbf{MoA} performs routed multi-expert cross-modal alignment and builds a modality memory; (ii) a \textbf{graph} models relational structure.
The pathway outputs are concatenated and fed to a classifier. Implementation details are deferred to Section~\ref{sec:method}.}
\label{fig:fig_2}
\end{figure*}
 
\section{Method}
\label{sec:method}

\subsection{Overview}
The complete architecture is illustrated in Figure~\ref{fig:fig_2}. Given text, audio, and video for each utterance (with conversational context), our approach centers learning on \emph{emotion hotspots}—short, high-intensity segments—while preserving helpful global context. We first apply HGF to integrate modality-wise hotspots with utterance-level global features, producing enhanced unimodal sequences. On top of the encoded unimodal representations (denoted later as $\{\mathbf{X}_m\}$), we use MoA to address cross-modal temporal offsets and a cross-modal graph to inject relation-aware dialogue structure. The two streams are concatenated and fed to a token-level classifier. This separation lets HGF surface the most informative local evidence, while MoA and the graph branch handle alignment and conversational dependencies.

\subsection{Inputs and Hotspot-Gated Fusion}
We use three modalities $m\!\in\!\{\mathrm{T},\mathrm{A},\mathrm{V}\}$. For each utterance $u$ and modality $m$, we prepare a global content sequence $\mathbf{C}_m$ and a hotspot sequence $\mathbf{H}_m$ that highlights localized emotional salience. For vision, we extract motion-sensitive regions using optical-flow-based features; for audio, we obtain prosodic bursts via emphaclass~\cite{de2023emphassess}; for text, we concatenate the conversational context and leverage a strong Large Language Model~\cite{liu2024deepseek} to select salient spans.

\textbf{Hotspot-Gated Fusion.}
HGF performs token-wise gated interpolation between $\mathbf{C}_m$ and $\mathbf{H}_m$:
\begin{equation}
\begin{aligned}
\alpha_m &= \sigma\!\big(W_m[\mathbf{C}_m\,\|\,\mathbf{H}_m]+b_m\big),\quad \alpha_m\!\in\!\mathbb{R}^{L\times 1},\\
\mathbf{Z}_m &= \mathbf{C}_m + \alpha_m \odot (\mathbf{H}_m - \mathbf{C}_m),
\end{aligned}
\end{equation}
where $L$ is sequence length and $\odot$ denotes broadcasted element-wise multiplication. A per-modality encoder $\mathrm{Enc}_m$ then maps $\mathbf{Z}_m$ to a shared hidden space:
$\mathbf{X}_m = \mathrm{Enc}_m(\mathbf{Z}_m).$

\subsection{MoA for Cross-modal Alignment}
\label{sec:moa}
We introduce MoA for each ordered modality pair $(j\!\leftarrow\!k)$ to flexibly select and fuse source information from modality $k$ for aligning the target modality $j$. For each $(j\!\leftarrow\!k)$ we instantiate $E$ lightweight aligner experts (a single cross-attention layer with a small feed-forward):
\begin{equation}
\begin{aligned}
f_{j\leftarrow k}^{(e)}:\ (\mathbf{X}_j[t],\mathbf{X}_k)\mapsto \mathbb{R}^{h},\quad e=1,\dots,E.
\end{aligned}
\end{equation}
We perform utterance-level routing: for each target utterance $t$, concatenate the target embedding with a mean-pooled source context and produce expert logits, 
\begin{equation}
\begin{aligned}
\boldsymbol{\ell}_t &=\mathrm{Router}\!\big([\mathbf{X}_j[t];\,\mathrm{MeanPool}(\mathbf{X}_k)]\big)\in\mathbb{R}^{E},\\
\boldsymbol{\pi}_t &=\mathrm{Softmax}\!\big(\mathrm{TopKMask}(\boldsymbol{\ell}_t,K)\big).
\end{aligned}
\end{equation}
TopK is implemented as a masked softmax; experts are densely evaluated and linearly mixed (gated sparsification). The expert mixture is fed into a shallow cross-attention block to restore capacity, yielding
\begin{equation}
\mathbf{H}_{j\leftarrow k} = \mathrm{CrossAttnBlock}\!\Big(\sum_{e=1}^{E}\pi_{t,e}\,f_{j\leftarrow k}^{(e)}(\mathbf{X}_j,\mathbf{X}_k),\ \mathbf{X}_k\Big).
\end{equation}
For each target modality $j$, we concatenate aligned representations from all sources and apply self-attention to build a modality memory, then concatenate across all targets to form the final cross-modal representation:
\begin{equation}
\begin{aligned}
\mathbf{H}^{\mathrm{mem}}_j&=\mathrm{SelfAttn}\!\Big(\mathrm{concat}_{k\neq j}\,[\mathbf{H}_{j\leftarrow k}]\Big),\\
\mathbf{H}^{\mathrm{moa}}&=\mathrm{concat}_{j}\,[\mathbf{H}^{\mathrm{mem}}_j].
\end{aligned}
\end{equation}
This retains utterance granularity and composes cross-modal context without length pooling.

\subsection{Cross-Modal Graph Pathway}
\label{sec:graph}
We build an utterance-level multi-relational graph whose nodes are (modality $m$, utterance $t$). Edges consist of intra-modal temporal links within a window and optional same-time cross-modal links; each edge is typed by its temporal direction $s\!\in\!\{-1,0,1\}$ and the source/target modality pair. A relation-aware GNN produces node embeddings that are then concatenated per utterance across modalities:
\begin{equation}
\mathbf{H}^{\mathrm{gnn}}
=\mathrm{MultiConcat}\!\big(\mathrm{GNN}(\mathbf{X}^{\mathrm{graph}},\mathcal{E}),\,\text{lengths},\,|\mathcal{M}|\big).
\end{equation}
Here $\mathbf{X}^{\mathrm{graph}}$ packs the encoded unimodal sequences as node features, $\mathcal{E}$ is the set of typed edges, $|\mathcal{M}|$ is the number of modalities, and $\mathrm{MultiConcat}(\cdot)$ restores utterance-wise, modality-wise concatenation. This pathway supplies structured conversational context complementary to MoA.

\subsection{Loss Function}
We optimize a sum of the utterance-level task loss and an optional MoA load-balancing regularizer:
\begin{equation}
    \mathcal{L} \;=\; \mathcal{L}_{\mathrm{task}} \;+\; \lambda\,\mathcal{L}_{\mathrm{lb}}.
\end{equation}
Here $\lambda\!\ge\!0$ is a scalar weight (we use $\lambda\!>\!0$ by default).

The task loss is the negative log-likelihood over utterances:
\begin{equation}
    \mathcal{L}_{\mathrm{task}} \;=\; -\,\mathbb{E}_{(b,t)}\big[\log p(y_{b,t}\mid \mathbf{H}_{b,t})\big],
\end{equation}
where $(b,t)$ indexes batch $b$ and utterance $t$, $\mathbf{H}_{b,t}$ is the model’s utterance representation (concatenating the MoA and graph pathways), and $p(\cdot\mid\mathbf{H}_{b,t})$ is produced by the classifier; class-balancing weights can be applied when needed.

The load-balancing term encourages uniform expert usage in MoA:
\begin{equation}
\begin{aligned}
     \mathcal{L}_{\mathrm{lb}} \;&=\; \sum_{e=1}^{E} u_e \log u_e \;+\; \log E,\\
u_e \;&=\; \mathbb{E}_{(b,t)}\!\big[\pi_{t,e}\big].   
\end{aligned}
\end{equation}
Here $E$ is the number of experts, $\pi_{t,e}$ is the Top\textit{K}-masked softmax routing weight, and $u_e$ is the average expert usage.

\begin{table*}[!t]
\centering
{\caption{Comparison with previous state-of-the-art methods on IEMOCAP (6-way). The \textbf{bolded} results indicate the best
performance, while the \underline{underlined} results represent the second best performance. \label{tab:6way}
}}
\setlength{\tabcolsep}{4pt}
\begin{tabular}{|l|c c c c c c|c c|}
\hline
Methods  & \multicolumn{8}{c|}{IEMOCAP(6-way) }\\
\cline{2-9}
& Happy & Sad  & Neutral & Angry & Excited & Frustrated & Acc.(\%) & w-F1(\%)  \\
\hline
DialogueRNN~\cite{DialogueRNN}   & 33.18 & 78.80 & 59.21 & 65.28 & 71.86 & 58.91 & 63.40 & 62.75 \\
DialogueGCN~\cite{DialogueGCN}  & 47.10 & 80.88 & 58.71 & 66.08 & 70.97 & 61.21 & 65.54 & 65.04 \\
DialogueCRN~\cite{DialogueCRN}  & 51.59 & 74.54 & 62.38 & 67.25 & 73.96 & 59.97 & 65.31 & 65.34 \\
MMGCN~\cite{MMGCN} & 45.45 & 77.53 & 61.99 & 66.70 & 72.04 & 64.12 & 65.56 & 65.71 \\
COGMEN~\cite{cogmen}  & 55.76 & 80.17 & 63.21 & 61.69 & 74.91 & 63.90 & 67.04 & 67.27 \\
CORECT~\cite{nguyen-etal-2023-conversation}  & 56.19 & 82.11 & 63.97 & 66.49 & 69.12 & 61.91 & 66.61 & 66.74 \\
M3NET~\cite{M3NET} & 56.77 & \underline{82.26} & 68.94 & 66.67 & 78.06 & 60.32 & 69.38 & 69.29 \\
CMERC~\cite{CMERC} & \textbf{60.73} & 81.89 & \underline{71.65} & \textbf{69.51} & 77.45 & \textbf{67.02} & - & \underline{71.98} \\
\hline
baseline & 56.88 & 79.19 & 64.34 & 64.62 & 70.59 & 63.23 & 66.67 & 66.84 \\
baseline + HGF & 56.69 & 80.33 & 70.48 & 68.91 & \underline{78.25} & 63.45 & \underline{70.67} & 70.36 \\
baseline + HGF + MoA(Ours) & \underline{58.97} & \textbf{83.67} & \textbf{72.35} & \underline{69.23} & \textbf{79.13} & \underline{66.94} & \textbf{72.52} & \textbf{72.52} \\ 
\hline
\end{tabular}
\end{table*}

\begin{table*}[!t]
\centering
{\caption{Results on IEMOCAP (4-way) and CMU-MOSEI dataset compared with previous works. For sentiment classification on the CMU-MOSEI dataset, we trained an independent binary classifier for each emotion. The \textbf{bolded} results indicate the best performance, while the \underline{underlined} results represent the second best performance. $^*$ denotes identical scores for Fear (84.47) and Surprise (85.66) due to the baselines’ binary classifiers failing to separate the positive and negative classes. \label{tab:4way_CMU}}}
\setlength{\tabcolsep}{4pt}
\begin{tabular}{|l|c c|c c c c c c|}
\hline
Methods  & \multicolumn{2}{c|}{IEMOCAP(4-way)} &  \multicolumn{6}{c|}{CMU-MOSEI (w-F1(\%))} \\
\cline{2-9}
& Acc.(\%) & w-F1(\%)  & Happiness & Sadness  & Angry & Fear & Disgust & Surprise  \\
\hline
bc-LSTM~\cite{bc-LSTM} & 75.20 & 75.13 & - & - & - & - & - & -  \\
CHFusion~\cite{CHFusion} & 76.59 & 76.80 & - & - & - & - & - & -  \\
Multilouge-Net~\cite{multilogue-net} & - & - & 67.84 & 65.34 & 67.03 & 84.47$^*$ & 74.91 & 85.66$^*$  \\
TBJE~\cite{TBJE} & - & - & 65.91 & 68.54 & 70.78 & 84.47$^*$ & 80.23 & 85.66$^*$  \\
COGMEN~\cite{cogmen} & 82.29 & 82.15 & 68.73 & 69.11 & 74.20 & 84.47$^*$ & 80.23 & 85.66$^*$  \\
CORECT~\cite{nguyen-etal-2023-conversation} & \underline{82.93} & \underline{82.92} & \underline{69.24} & \underline{70.67} & \underline{76.36} & 84.47$^*$ & \underline{82.84} & \textbf{86.00} \\

\hline

Ours & \textbf{85.90} & \textbf{85.89} & \textbf{71.04} & \textbf{72.34} & \textbf{76.43} & \textbf{84.81} & \textbf{85.21} & \underline{85.84} \\

\hline
\end{tabular}
\end{table*}
\section{Experiments}
\label{sec:expri}


\subsection{Experimental Setting}
\textbf{Dataset.}
We evaluate our method on two public multimodal ERC datasets: IEMOCAP~\cite{busso2008iemocap} and CMU-MOSEI~\cite{CMU-MOSEI}.
IEMOCAP contains 151 dialogues (7433 utterances, 12 hours in total) annotated with six emotions: happy, sad, neutral, angry, excited, and frustrated. To reduce label ambiguity, (happy, excited) and (sad, frustrated) are merged into a four-class setting.
CMU-MOSEI provides annotations for six basic emotions: happiness, sadness, anger, fear, surprise, and disgust.

\textbf{Evaluation Metrics}. The primary metrics for evaluation are the \emph{weighted F1-score} (w-F1) and \emph{Accuracy} (Acc.). The w-F1 metric is computed as the sum $\sum_{k=1}^{K} \text{freq}_{k} \times \text{F1}_{k}$, where $\text{freq}_k$ is the relative frequency associated with class \emph{k}. Accuracy, conversely, measures the percentage of correct predictions over the entire test set.


\textbf{Baselines and Implementation.} 
Our model extends the CORECT framework~\cite{nguyen-etal-2023-conversation} with modifications while ensuring fair comparison with baseline methods. On IEMOCAP, we conduct both six-class and four-class experiments, while on CMU-MOSEI, we perform binary classification for each emotion category.
The implementation is based on PyTorch 2.1.1 and torch-geometric, trained on 4×L20 GPUs. We fix the batch size and random seed across all experiments. Adam optimizer is used, with a ReduceLR scheduler to decrease the learning rate when validation accuracy plateaus.





\subsection{Overall Results}
We compare our model against dataset-specific SOTA baselines, which include RNN-based models and graph-based methods.
Tables~\ref{tab:6way} and \ref{tab:4way_CMU} show that augmenting the baseline graph pathway with HGF and the MoA yields consistent, architecture-driven gains. HGF injects an emotion hotspot prior at the unimodal level, strengthening emotion-sensitive cues without length pooling and bringing steadier class-wise improvements—particularly on noise-prone categories such as Neutral and Excited. MoA then performs target-specific cross-modal alignment via TopK-masked expert routing followed by a lightweight capacity restoration block, mitigating confusions among semantically similar emotions (e.g., Happy/Excited and Sad/Frustrated). 

Running HGF and MoA in parallel with the graph pathway and concatenating their outputs at the utterance level leads to state-of-the-art results, including 72.52 Acc./w-F1 on IEMOCAP (6-way) and 85.90/85.89 on IEMOCAP (4-way), with leading scores on most categories in CMU-MOSEI; the tables provide the detailed per-class trends that underpin these gains.


\subsection{Ablation Study}
As Tabel~\ref{tab:6way} shows, we ablate modules incrementally atop the original baseline, which already includes the cross-modal graph pathway. We keep the graph pathway intact for fairness to baseline~\cite{nguyen-etal-2023-conversation} and because our design complements—rather than replaces—its relational/temporal modeling. 

Adding HGF provides the first substantial improvement by amplifying emotion hotspot-guided local cues and reducing reliance on aggressive pooling, with table evidence of stronger Neutral/Excited performance and overall stability. Adding MoA on top delivers further, steady gains by selectively routing cross-modal evidence to the target utterance and sharpening alignment where it is most discriminative, reflected in the consistent transition from baseline to +HGF to +HGF+MoA in both aggregate and class-wise results.

\section{Conclusion}
\label{sec:conclu}
We define \emph{emotion hotspots}—short, high-intensity bursts within each modality—and place them at the center of ERC. Yet hotspots across modalities seldom peak simultaneously, making alignment essential. To resolve cross-modal misalignment while preserving context, we couple an adaptive Hotspot-Gated Fusion (HGF) for hotspot–global integration with a routed Mixture-of-Aligners (MoA) cross-attention for alignment, and add a conversational graph pathway to encode dialogue structure. The resulting model attains state-of-the-art results on IEMOCAP (6-way and 4-way) and ranks best on most emotions in CMU-MOSEI while remaining competitive elsewhere. Ablations show that HGF yields substantial gains that are further amplified by MoA—with the largest improvements on \textit{Excited} and \textit{Neutral}—and expert routing offers interpretability, underscoring a robust, general approach to multimodal ERC.


\newpage
\newpage
\ninept
\section*{Acknowledgments}
This work was supported in part by the Scientific Research Starting Foundation of Hangzhou Institute for Advanced Study (2024HIASC2001), the Zhejiang Provincial Natural Science Foundation of China (No.\ LQN25F020001), and in part by the Key R\&D Program of Zhejiang (2025C01104).

\noindent\textbf{Use of Generative AI and AI-Assisted Tools.}
Language editing in throughout the manuscript was assisted by ChatGPT (OpenAI) to improve grammar and clarity; all scientific content was authored by the authors.
During implementation, the authors used Cursor (an AI code assistant) for debugging support; no AI-generated code, figures, tables, or text were included in the manuscript.
All AI-assisted outputs were reviewed and verified by the authors, who take full responsibility for the content.

\section*{Compliance with Ethical Standards}
This study involved no human or animal subjects and did not require ethics approval.

\end{document}